\documentclass[10pt,conference]{style_file}
\usepackage{graphicx}
\usepackage{booktabs}
\usepackage{amssymb}
\usepackage{amsmath}
\usepackage{algorithm}
\usepackage{algorithmic}

\newcommand{\comment}[1]{}
\newtheorem{theorem}{Theorem}

\begin{document}

\title{Belief Flows for Robust Online Learning}

\author{\authorblockN{Pedro~A.~Ortega}
\authorblockA{School of Engineering\\ and Applied Sciences\\
University of Pennsylvania\\
Philadelphia, PA 19104, USA\\
Email: ope@seas.upenn.edu}
\and
\authorblockN{Koby~Crammer}
\authorblockA{Department of\\
Electrical Engineering\\
The Technion\\
Haifa, 32000 Israel\\
Email: koby@ee.technion.ac.il}
\and
\authorblockN{Daniel~D.~Lee}
\authorblockA{School of Engineering\\ and Applied Sciences\\
University of Pennsylvania\\
Philadelphia, PA 19104, USA\\
Email: ddlee@seas.upenn.edu}}
%

\maketitle

\begin{abstract}
  This paper introduces a new probabilistic model for online learning
  which dynamically incorporates information from stochastic gradients
  of an arbitrary loss function.  Similar to probabilistic filtering,
  the model maintains a Gaussian belief over the optimal weight
  parameters.  Unlike traditional Bayesian updates, the model
  incorporates a small number of gradient evaluations at locations
  chosen using Thompson sampling, making it computationally tractable.
  The belief is then transformed via a linear flow field which
  optimally updates the belief distribution using rules derived from
  information theoretic principles.  Several versions of the algorithm
  are shown using different constraints on the flow field and compared
  with conventional online learning algorithms.  Results are given for
  several classification tasks including logistic regression and
  multilayer neural networks.
\end{abstract}

\section{Introduction}\label{sec:introduction}

An number of problems in artificial intelligence must cope with
continual learning tasks involving very large datasets or even data
streams which have become ubiquitous in application domains such as
life-long learning, computer vision, natural language processing,
bioinformatics and robotics. As these big-data problems demand richer
models, novel online algorithms are needed that scale efficiently both
in accuracy and computational resources. Recent work have shown that
complex models require regularization to avoid local minima and
overfitting---even when the data is abundant
\cite{Welling2011}.

The aim of our work is to formulate an efficient online learning
approach that leverages the advantages of \emph{stochastic gradient
  descent} (SGD) \cite{RobbinsMonro1951} and \emph{Bayesian filtering}
\cite{Sarkka2013}.  Many learning tasks can be cast as optimization
problems, and SGD is simple, scalable and enjoys strong theoretical
guarantees in convex problems \cite{Bottou2010,Bach2011} but tends to
overfit if not properly regularized. On the other hand, Bayesian
filtering methods track belief distributions over the optimal
parameters to avoid overfitting, but are typically computationally
prohibitive for rich models.  To combine these two approaches, we had
to address two questions.

\begin{figure}
\centering
\includegraphics[width=\columnwidth]{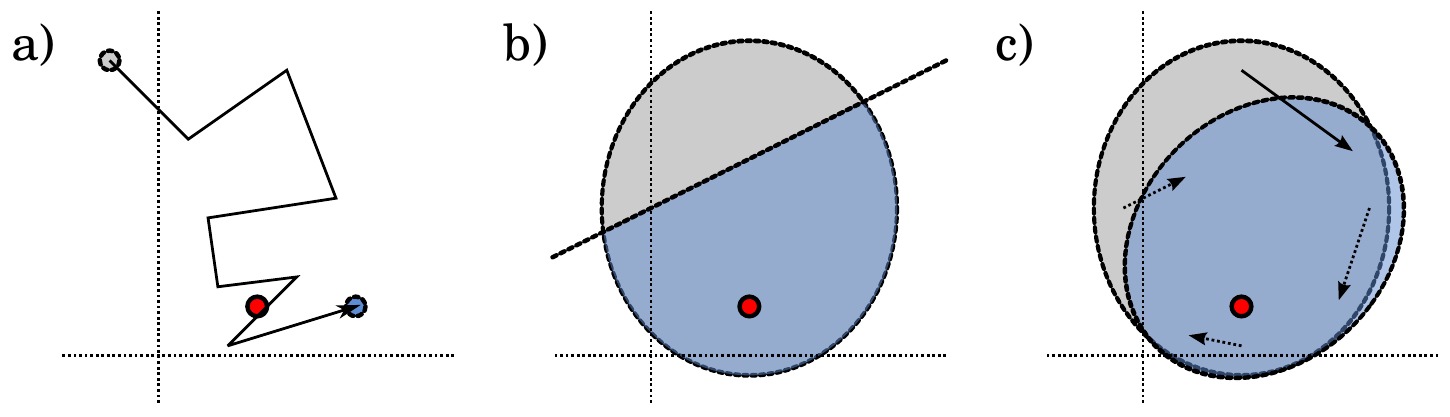}
\caption{Schematic comparison of learning dynamics in (a) stochastic
  gradient descent, (b) Bayesian filtering, and (c) belief flows.}
\label{fig:bflo_intro}
\vspace{-15pt}
\end{figure}

The first question is how to update a global belief distribution over optimal 
parameters from a local update prescribed by SGD. In other words,
rather than calculating the posterior using a likelihood function,
we take gradients as \emph{directly} specifying the velocity field---or 
\emph{flow field}---of belief updates. As shown in Figure 
\ref{fig:bflo_intro}, this can be viewed as tracking an ensemble of 
models under the dynamics induced by the expected prediction error.  
We answer this question by using the principle of \emph{minimum information 
discrimination} (MID) \cite{Kullback1959} to choose the most conservative
posterior belief that is consistent with the gradient measurement 
and assumptions on the flow field. 

The second question is how to generate predictions without integrating 
over the parameter uncertainty. Calculating the optimal belief update 
would require estimating the expected SGD update at every point in parameter
space, which is prohibitive. To overcome this problem, a natural choice
is to use \emph{Thompson sampling}, i.e. sampling parameters from the 
posterior according to the probability of being optimal \cite{Thompson1933}. 
As is well known in the literature in Bayesian optimization \cite{Mockus2013}, 
optimizing the parameters  of an unknown and possibly non-convex error 
function requires dealing with the exploration-exploitation trade-off, 
which is precisely where Thompson sampling has been shown to outperform
most state-of-the-art methods \cite{Chapelle2011}.

Here, we illustrate this modelling approach by deriving the update rules for 
Gaussian belief distributions over the optimal parameter. By assuming that 
observations generate linear flow fields, we furthermore show that the
resulting update rules have closed-form solutions. Because of this, the
resulting learning algorithms are online, permitting training examples
to be discarded once they have been used.

\section{Gaussian Belief Flows}

\comment{
\begin{enumerate}
\item Explain the method.
\item Derive rules for Gaussian beliefs with linear flows.
\item Present pseudocode.
\end{enumerate}
}

We focus on prediction tasks with parameterized models, each denoted by
$F_w(x)$ with inputs $x \in \mathbb{R}^p$ and parameters $w \in
\mathbb{R}^d$. At each round $n=1,2,\ldots$ the algorithm maintains a belief
distribution $P_n(w)$ over the optimal parameters. Here we choose $P_{n}(w)$ to
be represented by a $d$-dimensional Gaussian with mean $\mu_{n}$ and covariance
$\Sigma_{n}$:
\begin{align*}
&P_{n}(w) = N(w; \mu_{n},\Sigma_{n}) \\
&= \frac{1}{\sqrt{(2\pi)^d \,{\rm det}\Sigma_n}}
  \exp\left[-\frac{1}{2} (w-\mu_{n})^{T}\Sigma_{n}^{-1}(w-\mu_{n})\right].
\end{align*}
In each round, the algorithm samples a parameter vector $w_{n}$ from
the distribution $P_{n}(w)$. The components of $w_{n}$ are then used as
parameters in a classification or regression machine for a given input $x_{n}$.
For example, we can consider logistic regression where $w_{n}$
are the parameters of the logistic function. Or we can use deep networks, where
$w_{n}$ specifies the synaptic weights of the different layers of the network.
Without loss of generality, this yields a prediction $\hat{y}_n$ for the
particular input~$x_{n}$:
\[
  \hat{y}_n = F_{w_{n}}(x_{n}).
\]
A supervised output signal $y_{n}$ is also provided, and we wish
to minimize the loss between the true output and predicted output:
\[
  \ell(y_{n}, \hat{y}_n).
\]
To improve the loss on the current example, SGD then updates the parameter as:
\begin{equation}\label{eq:sgd}
  w_{n}^{\prime} = w_n - \eta \cdot
    \frac{\partial}{\partial w}\ell(y_{n},\hat{y}_n)\vert_{w=w_{n}},
\end{equation}
where $\eta>0$ is a learning rate which may vary as the number of rounds
increase. For the case of a multilayer perceptron, this gradient can
be efficiently computed by the well-known backpropagation algorithm
in a single backwards pass.

\comment{
\begin{figure}[tbp]
\centering
\includegraphics[width=\textwidth]{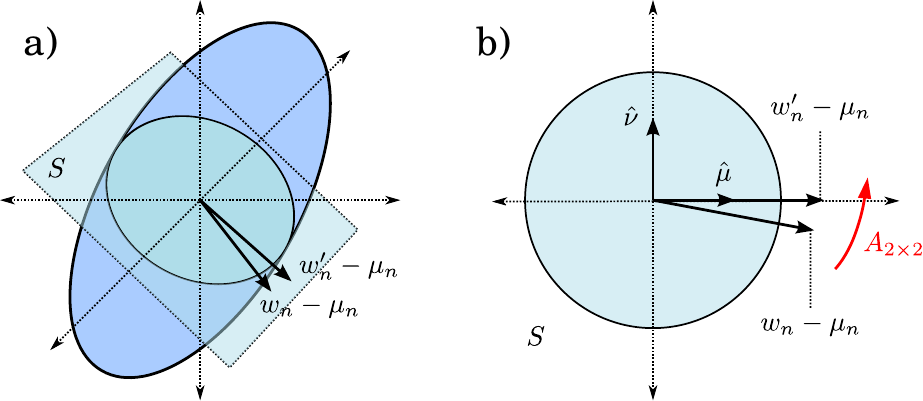}
\caption{The belief flow update. (a) The update from the prior (shaded) to the
posterior (solid) minimizes the KL divergence subject to the gradient
descent update $w_n \rightarrow w_n'$ of a randomly drawn point. (b) $S$ is the
2-D linear subspace spanned by $(w_n-\mu_n)$ and $(w'_n-\mu_n)$. (c)
Illustration of the linear transformation $A_{2 \times 2}$ in the linear
subspace $S$.}
\label{fig:illustration}
\end{figure}
}

\subsection{Full Flow}

Knowing that the sampled parameter vector $w_n$ needs to be modified to
$w^\prime_n$, how do we choose the posterior $P_{n+1}(w)$? To answer this
question, we assume that the update results from a linear flow field,
i.e.\ each $w \in \mathbb{R}^d$ is updated as
\[
  w^\prime = Aw + b
\]
where $A \in \mathbb{R}^{d \times d}$ is an affine transformation
matrix and $b \in \mathbb{R}^d$ is a translational offset.  The main
advantage of a linear flow is that Gaussian distributions remain
Gaussian under such a field \cite{Crammer2010}. In particular, we
choose the flow parameters to match the SGD update~\eqref{eq:sgd}:
\begin{equation}\label{eq:weight_constraint}
  w_n^{\prime} = A w_n + b.
\end{equation}
Under this linear flow, the new belief distribution is
\[
  P_{n+1}(w) = N(A\mu_{n}+b,\, A\Sigma_{n}A^{T})=N(\mu_{n+1},\,\Sigma_{n+1})
\]
where the mean is shifted to $\mu_{n+1} = A \mu_{n} + b$ and the
covariance is scaled to $\Sigma_{n+1} = A \Sigma_{n} A^{T}$. There are
many potential $A$ and $b$ which satisfy the flow constraint in
\eqref{eq:weight_constraint}, so we need a way to regularize the
ensuing distribution $P_{n+1}(w)$.  We utilize the Kullback-Leibler
divergence
\begin{equation}\label{eq:min-kl}
  \min_{A,b} D_{\rm KL}\left[P_{n+1} \| P_n \right]
  = \min_{A,b} \int P_{n+1}(w)\log\frac{P_{n+1}(w)}{P_{n}(w)}dw.
\end{equation}
subject to the constraint $w_n^{\prime} = Aw_n + b$. This is an
application of the principle of MID \cite{Kullback1959}, an extension
of the maximum entropy principle \cite{Jaynes1957a}, and governs the
updates of exponential family distributions \cite{Csiszar2004}. The
next theorem (see Appendix for the proof) shows that this 
problem has a closed-form solution.

\begin{theorem}\label{theo:full-solution}
Let $\Sigma_n = U_n D_n U_n^T$ be the eigendecomposition of the covariance
matrix. Let
\begin{align*}
  u \hat{\mu}
  &= \frac{1}{\sqrt{D_n}} U^T_n (w_n-\mu_n) \\
  v_{\parallel} \hat{\mu} + v_{\perp} \hat{\nu}
  &= \frac{1}{\sqrt{D_n}} U^T_n (w^\prime_n-\mu_n)
\end{align*}
be the transformed (whitened) differences between the sampled parameter vector
and the mean before and after the update respectively, expressed in terms of the
2-D basis spanned by the unitary orthogonal vectors $\hat{\mu}$ and $\hat{\nu}$.
Then, the solution $A^\ast$ to \eqref{eq:min-kl} is
\[
  I_{d \times d} + U_n \sqrt{D_n}
  \left\{
  \bigl[\hat{\mu} \hspace{6pt} \hat{\nu}\bigr]
  \left(A_{2\times2} - I_{2\times2}\right)
  \left[\begin{array}{c}
  \hat{\mu}^{T}\\
  \hat{\nu}^{T}
  \end{array}\right]
  \right\}
  \frac{1}{\sqrt{D_n}} U^T_n
\]
where the 2-D transformation matrix $A_{2 \times 2}$ is given by
\[
\frac{1}{\sqrt{v_{\parallel}^{2}+v_{\perp}^{2}}}
\left[\begin{array} {cc}
\frac{u\sqrt{v_{\parallel}^{2}+v_{\perp}^{2}}
  + \delta_1 \sqrt{4+u^{2}(4+v_{\parallel}^{2}
  + v_{\perp}^{2})}}{2(1+u^{2})}v_{\parallel}
& - \delta_2 v_{\perp} \\
\frac{u\sqrt{v_{\parallel}^{2} + v_{\perp}^{2}}
  + \delta_1 \sqrt{4+u^{2}(4+v_{\parallel}^{2}
  + v_{\perp}^{2})}}{2(1+u^{2})}v_{\perp}
& + \delta_2 v_{\parallel}
\end{array}\right]
\]
$\delta_1, \delta_2 \in \{-1,+1\}$. The hyperparameters of the 
posterior distribution are then equal to
\[
\Sigma_{n+1}=A^\ast \Sigma_{n} A^{\ast T}
\qquad
\mu_{n+1}=A^\ast(\mu_{n}-w_{n})+w_{n}^{\prime}.
\]
\end{theorem}

There are actually four discrete solution to \eqref{eq:min-kl}, one for
each combination of $\delta_1$ and $\delta_2$. If we assume that the
SGD learning rate is small, then $w'_n-\mu_n \approx w_n-\mu_n$. In this case,
we should choose the solution $A^\ast \approx I_{d \times d}$ obtained by
selecting $\delta_1 = \delta_2 = 1$. 

\comment{
\begin{figure}
\centering
\includegraphics[width=0.5\textwidth]{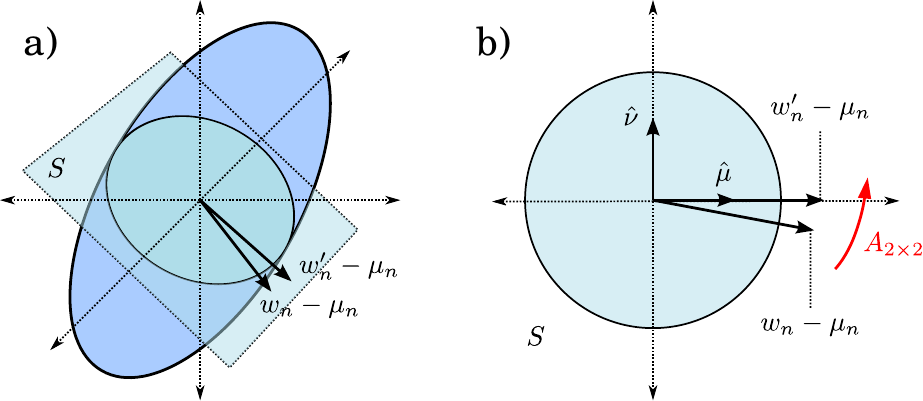}
\caption{2-D Subspace of the Transformation}
\label{fig:illustration}
\end{figure}
The proof of Theorem~\ref{theo:full-solution} can be found in the appendix.
Intuitively, the first step points out that the optimal transformation matrix
$A^\ast$ essentially acts on the $2$-D space spanned by $(w_n-\mu_n)$ and
$(w'_n-\mu_n)$; it is identity in the other $D-2$ directions
orthogonal to this 2-D subspace (Figure~\ref{fig:illustration}a). The second
step calculates the optimal transformation matrix $A_{2 \times 2}$ for this 2-D
subspace (Figure~\ref{fig:illustration}b).
}

\subsection{Diagonal and Spherical Flows}

We now consider two special cases: flows with diagonal and spherical
transformation matrices, i.e. of the form
\[
  A = \mathrm{diag}(a_1, a_2, \ldots, a_d)
  \quad\text{and}\quad
  A = a U,
\]
where $\mathrm{diag}(v)$ denotes the square diagonal matrix with the elements of
the vector $v$ on the main diagonal and $U$ is a unitary (rotation) matrix. We
match these flow types with multivariate Gaussians having diagonal and
spherical covariance matrices. Let $N(\mu,\Sigma)$ be the prior and
$N(\mu',\Sigma')$ be the posterior at round $n$, and let subindices denote
vector components in the following.

\paragraph{Diagonal:}
For the diagonal case, the multivariate
distribution $N(\mu,\Sigma)$ factorizes into $d$ univariate Gaussians $N(\mu_i,
\sigma_i^2)$ that can be updated independently under flow fields of the form
\begin{equation}\label{eq:transformation_diagonal}
  w_i^{\prime} = a_i w_i + b_i.
\end{equation}
Under these constraints, it can be shown (see Appendix) that the optimal
transformation is given by
\[\label{eq:solution_diagonal}
  a_i^\ast = \frac{u_i v_i + \delta_i \sqrt{4+u_i^2(4+v_i^2)}}{2(1+u_i^2)},
\]
where $u_i = (w_i-\mu_{i})/\sigma_i$, $v_i = (w_i^{\prime}-\mu_i)/\sigma_i$
are the $i$-th component of the normalized sampled parameter before and after
updating and $\delta_i \in \{-1,+1\}$. Similarly to the general case, we choose
the solution closer to the identity when $u_i \approx v_i$ by picking
$\delta_i = +1$. The posterior hyperparameters are
\begin{equation}\label{eq:update_diagonal}
  \sigma_i^\prime = a_i^\ast \sigma_i
  \qquad
  \mu_i^\prime = a_i^\ast (\mu_i-w_i) + w_i^\prime.
\end{equation}
where $\mu_i^\prime$ and $\sigma_i^\prime$ are the new mean and standard
deviation of the $i$-th component.

\paragraph{Spherical:}
For the spherical case, the flow field that preserves the spherical
distribution is of the form $A = a U$, where $a > 0$ is a scalar and $U$ is a
unitary matrix such that $A(w-\mu)$ and $(w'-\mu)$ are colinear; that is, $A$
rotates and scales $(w-\mu)$ to align it to $(w'-\mu)$. The update is
\begin{equation}\label{eq:transformation_spherical}
  w' = a w + b,
\end{equation}
that is, similar to \eqref{eq:transformation_diagonal} but with an
isotropic scaling factor $a$. The optimal scaling factor is then
given by
\[
  a^\ast = \frac{u v + \delta \sqrt{4+u^2(4+v^2)}}{2(1+u^2)},
\]
where $u = (\| w-\mu \|)/\sigma$, $v = (\| w^{\prime}-\mu \|)/\sigma$,
where $\delta \in \{-1, +1\}$ is choosen as $\delta = 1$ for the near-identity
transformation.

\subsection{Non-Expansive Flows}

The previously derived update rules allow flow fields to be expansive,
producing posterior distributions having larger differential entropy
than the prior. Such flow fields are needed when the error landscape
is dynamic, e.g.\ when the data is nonstationary. However, for faster
convergence with stationary distributions, it is desirable to restrict
updates to non-expansive flows. Such flow fields are obtained by
limiting the singular values of the transformation matrix $A$ to
values that are smaller or equal than one.

\subsection{Implementation}

The pseudocode of a typical gradient-based online learning
procedure is listed in Algorithm~\ref{alg:bflo}. For a
$d$-dimensional multivariate Gaussian with diagonal and spherical covariance
matrix, the update has time complexity $O(d)$, while for an unconstrained
covariance matrix, this update is $O(d^3)$ in a naive implementation performing
a spectral decomposition in each iteration. The complexity of the unconstrained
covariance implementation can be reduced using low-rank techniques as
described in~\cite{bunch1978rank}. Numerically, it is important to maintain the
positive semidefiniteness of the covariance matrix. One simple way to achieve
this is by constraining its eigenvalues to be larger than a predefined minimum.
\begin{algorithm}[htb]
\caption{BFLO Pseudo-Code}
\label{alg:bflo}
\centering
\begin{algorithmic}
  \STATE {\bf Input:} $\mu_1, \Sigma_1$, hyperparameters
  \FOR{ $n = 1,2,\ldots,N$ }
  \STATE Get training example $(x_n,y_n)$.
  \STATE \emph{Calculate output:}
  \STATE Sample $w_n \sim N(\mu_n, \Sigma_n)$
  \STATE Set $z_n \leftarrow F_w(x_n)$
  \STATE \emph{Local flow:}
  \STATE Calculate new weights using e.g. gradient descent,
  \\ $w'_n \leftarrow w_n
    - \eta \frac{\partial \ell}{\partial w}(y_n,z_n)|_{w_n}$
  \STATE \emph{Global flow:}
  \STATE Calculate flow matrix $A^\star$ (full, diagonal or spherical).
  \STATE \emph{Update hyperparameters:}
  \STATE $\Sigma_{n+1} \leftarrow A^\star \Sigma_n A^{\star T}$
  \STATE $\mu_{n+1} \leftarrow A^\star (\mu_n-w_n) + w'_n$
  \STATE (Optional) perform numerical correction.
  \ENDFOR
  \STATE {\bf Return} $(\mu_N, \Sigma_N)$
\end{algorithmic}
\end{algorithm}

\section{Properties}

\subsection{Comparison}

\begin{figure*}
\includegraphics[width=\textwidth]{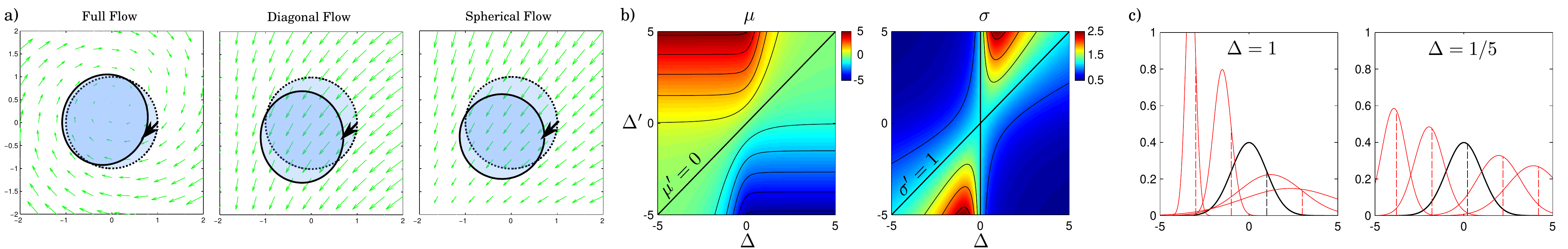}
\caption{a) Comparison of different flow types. b) 1-D Update of the mean
$\mu=0$ and standard deviation $\sigma=1$ as a function of $\Delta=(w-\mu)$
and $\Delta'=(w'-\mu)$. b) The two panels illustrate the posterior beliefs (red)
resulting from updating a 1-D prior (black) by moving a sampled weight through
displacements in $\{-4,-2,+2,+4\}$. The prior and posterior positions of the
sampled weight are indicated with dashed vertical lines.}
\label{fig:properties}
\end{figure*}

The full optimal transformation will include rotations in the 2-D
subspace spanned by the sampled and learned parameter vectors.  In
contrast, the diagonal transformation will only scale and translate
each basis direction independently, and the spherical transformation
acts on the radial parameter only. Since rotations allow for more
flexible transformations, the full flow update will keep the belief
distribution relatively unchanged, while the diagonal and spherical
flows will force the belief distribution to shift and compress
more. The update that is better at converging to an optimal set of
weights is problem dependent. The three update types are shown in 
Figure~\ref{fig:properties}a.

\subsection{Update Rule}\label{sec:update-rule}

To strenghten the intuition, we briefly illustrate the non-trivial effect
that the update rule has on the belief distribution.
Figure~\ref{fig:properties}b shows the values of the posterior
hyperparameters of a one-dimensional standard Gaussian as a function of $\Delta
= (w-\mu)$ and $\Delta' = (w'-\mu)$, that is, the difference of the sampled
weight and the center of the prior Gaussian before and after the update.

Roughly, there are three regimes, which depend on the displacement
$\Delta'-\Delta=w'-w$. First, when the displacement moves towards the prior mean
without crossing it, then the variance decreases. This occurs when $0 \leq
\Delta'\Delta \leq \Delta^2$, that is, below the diagonal in the first quadrant
and above the diagonal in the third quadrant. Second, if the displacement
crosses the prior mean, then the flow field is mainly explained in terms of a
linear translation of the mean. This corresponds to the second and fourth
quadrants. Finally, when the displacement moves away from the prior mean,
the posterior mean follows the flow and the variance increases. This corresponds
to the regions where $\Delta^{2} < \Delta'\Delta$, i.e.\ above the diagonal in
the first and below the diagonal in the third quadrant.

Note that the diagonal $\Delta'=\Delta$ leaves the two hyperparameters
unchanged, and that the vertical $\Delta = 0$ does not lead to a change of the
variance. Figure~\ref{fig:properties}c illustrates the change of the prior
into a posterior belief. The left panel shows the update of a sampled
weight equal to the standard deviation, and the right panel show the update for
a sampled weight equal to one-fifth of the standard deviation. Here, it is seen
that if the sampled weight is closer to the mean, then the update is
reflected in a mean shift with less change in the variance.

\subsection{Pseudo Datasets}

\begin{figure}
\centering
\includegraphics[width=0.9\columnwidth]{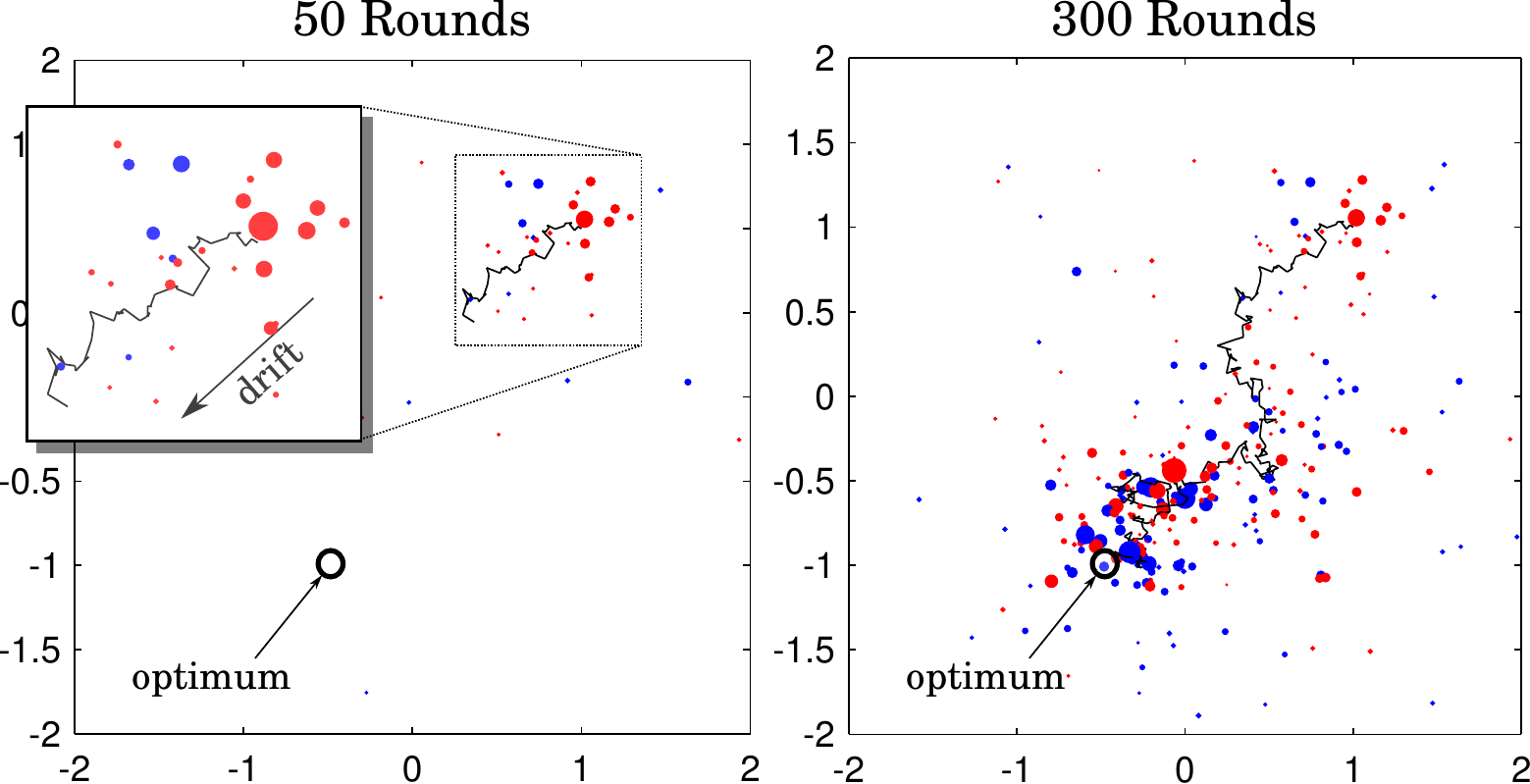}
\caption{Evolution of Pseudo Dataset}
\label{fig:pseudo-dataset}
\end{figure}

The belief updates can be related to Bayesian updates through (pseudo)
datapoints that would yield the same posterior. Since a belief flow update
ensures that the posterior stays within the Gaussian family, it is  natural to
relate it to Bayesian estimation of an unknown mean (but known covariance)
under the \emph{self-conjugate} Gaussian family. More precisely, let
$P(w)=N(w;\mu,\Sigma)$ and $P'(w)=N(w;\mu',\Sigma')$ denote the prior and  the
posterior of a belief flow update. Then, it is easily verified that this is
equivalent to conditioning a prior $P(w) = N(w;\mu,\Sigma)$ on a point $u \in
\mathbb{R}^d$ with likelihood function $P(x|w) = N(x;w,R)$, where
\begin{align}\label{eq:psuedo-datapoint}
  x &= (\Sigma^{\prime -1}-\Sigma^{-1})^{-1}
    (\Sigma'^{-1} \mu' - \Sigma^{-1} \mu), \\
  R &= (\Sigma^{\prime -1}-\Sigma^{-1})^{-1}.
\end{align}
The matrix $R$ is symmetric but not necessarily positive semidefinite unless we
use non-expansive flows. A negative eigenvalue $\lambda$ of $R$ then indicates
an \emph{increase} of the variance along the direction of its eigenvector
$v_\lambda \in \mathbb{R}^d$. From a Bayesian point of view, this implies that
the pseudo datapoint was removed (or forgotten), rather than added, along the
direction of $v_\lambda$. We have already encountered this case in the 1-D case
in Section~\ref{sec:update-rule} when the posterior variance increases due to a
displacement that points away from the mean.

Figure~\ref{fig:pseudo-dataset} shows the pseudo dataset for a sequence of
updates of a spherical belief flow. The temporal dynamics of the belief
distribution can be thought of as driven by the addition (subtraction) of
datapoints that attract (repel) the mean shown in black by relocating the center
of mass. In the figure, blue and red correspond to added and subtracted
datapoints respectively, and the circular areas indicate their precision, i.e.
$\rho_n = 1/\lambda_n$, where $\lambda_n \in \mathbb{R}$ is the unique
eigenvalue of $R$ at round $n$. The convergence of the belief distribution to a
particular point $z \in \mathbb{R}^d$ can thus be analyzed in terms of the
convergence of the weighted average of the pseudo datapoints to $z$ and
accumulation of precision, that is $\sum_n \rho_n \rightarrow +\infty$.

\section{Empirical Evaluation}

We evaluated the diagonal variant of Gaussian belief flows (BFLO) on a variety 
of classification tasks by training logistic regressors and multilayer 
neural networks. These results were compared to several baseline methods, 
most importantly stochastic gradient descent (SGD). Our focus in these
experiment was not to show better performance to existing learning 
approaches, but rather to illustrate the effect of different regularization 
schemes over SGD. Specifically, we were interested in the transient 
and steady-state regimes of the classifiers to measure the online 
and generalization properties respectively. 


\subsection{Logistic Regression}

For this model, we compared Gaussian belief flows (BFLO) on several 
binary classification datasets and compared its performance to 
three learning algorithms: AROW \cite{Crammer2009}, stochastic gradient 
descent (SGD) and Bayesian Langevin dynamics (BLANG) \cite{Welling2011}. 
With the exception of AROW, which combines large margin training and 
confidence weighting, these are all gradient-based learning algorithms. 
The algorithms were used to train a logistic regressor described as follows. 
The probability of the output $y \in \{0,1\}$ given the corresponding 
input $x \in \mathbb{R}^2$ is modelled as
\begin{equation} 
  \label{eq:logistic}
  P(y=1|x) = \sigma(w^T x), 
\end{equation}
where $w \in \mathbb{R}^2$ is a parameter
vector and $\sigma(t) = \frac{1}{1+\exp(-t)}$ is the logistic sigmoid. When
used in combination with the binary KL-divergence\footnote{Equivalently, one
can use the binary cross-entropy defined as $\ell(y,z) = -y\log z -(1-y)\log(1-z)$.} 
\begin{equation}\label{eq:binary-KL}
  \ell(z,y) = y\log\frac{y}{z} + (1-y)\log\frac{(1-y)}{(1-z)}
\end{equation}
as the error function, the error gradients become:
\begin{equation}\label{eq:cross-entropy-gradients}
  \frac{\partial\ell}{\partial w}
  = (z-y)x.
\end{equation}

We measured the online and generalization performance
in terms of the number of mistakes made in a single pass through 80\%
of the data and the average classification error without updating on the
remaining 20\% respectively. Additionally, to test robustness to noise, we
repeated the experiments, but inverting the labels on 20\% of the training
examples (the evaluation was still done against the true labels).

To test the performance in the online setting, we selected well-known binary
classification datasets having a large number of instances, summarized
as follows. {\bf MUSHROOM}: Physical characteristics of mushrooms, to be
classified into edible or poisonous. This UCI dataset contains 8124 instances
with 22 categorical attributes each that have been expanded to a total of 112
binary features\footnote{http://www.csie.ntu.edu.tw/~cjlin/libsvmtools/datasets/binary.html}. {\bf
COVTYPE}: 581,012 forest instances described by 54 cartographic
variables that have to be classified into 2 groups of
forest cover types \cite{Collobert2002}. {\bf IJCNN}: This is the
first task of the IJCNN 2001 Challenge \cite{Prokhorov2001}. We took the
winner's preprocessed dataset \cite{Chang2001}, and balanced the classes by
using only a subset of 27,130 instances of 22 features. {\bf EEG}: This
nonstationary time series contains a recording of 14,980 samples of 14 EEG
channels. The task is to discriminate between the eye-open and eye-closed
state\footnote{https://archive.ics.uci.edu/ml/datasets/EEG+Eye+State}. {\bf A9A}: This
dataset, derived from \emph{UCI Adult} \cite{Lin2008}, consists of 32,561
datapoints with 123 features from census data where the aim is to predict
whether the income exceeds a given threshold.

\begin{table*}
\newcommand{\kk}{\mathrm{K}}
\centering
\caption{Binary Classification Results for Noise Levels $0\%/20\%$}
\label{tab:results_logistic}
\scriptsize
\begin{tabular}{lcccccc}
\toprule
& \multicolumn{6}{c}{Online Classification Error in \%} \\
\midrule
& MUSHR. & COVTYPE & IJCNN & EEG & A9A & Rank \\
\midrule
AROW
&$\mathbf{5.32}/\mathbf{11.05}$ &$\mathbf{22.58}/\mathbf{23.10}$
&$\mathbf{8.44}/17.38$ &$43.59/45.11$ &$\mathbf{17.79}/20.95$
&$\mathbf{1.2}/2.8$\\
SGD
&$11.86/11.05$ &$28.03/28.18$ &$9.01/\mathbf{9.87}$ &$\mathbf{43.39}/44.22$
&$18.62/\mathbf{18.48}$
&$1.8/\mathbf{1.6}$\\
BLANG
&$14.44/14.35$ &$29.30/29.44$ &$12.86/12.06$ &$43.71/\mathbf{44.20}$
&$20.51/20.20$
&$3.2/2.8$\\
BFLO
&$14.30/15.34$ &$28.14/28.39$ &$10.34/11.52$ &$44.07/44.37$ &$19.03/19.04$
&$3.8/2.8$\\
\midrule
& \multicolumn{6}{c}{Final Classification Error in \%} \\
\midrule
& MUSHR. & COVTYPE & IJCNN & EEG & A9A & Rank \\
$\max\{\sigma_\mathrm{err}\}$ & $0.23/0.32$ & $0.12/0.40$ & $0.26/1.23$ &
$0.69/1.26$ & $0.08/0.12$ \\
\midrule
AROW
&$9.59/13.77$ &$37.18/38.11$ &$20.10/20.28$ &$65.38/63.09$
&$\mathbf{15.85}/17.56$ & $3.0/2.8$ \\
SGD
&$5.35/10.78$ &$37.45/38.36$ &$19.10/26.52$ &$\mathbf{60.57}/\mathbf{61.24}$ &
$17.45/19.32$ & $2.6/2.8$ \\
BLANG & $\mathbf{1.16}/\mathbf{0.34}$ & $38.39/39.09$ &
$\mathbf{15.97}/\mathbf{14.60}$
  & $64.85/66.68$ & $17.68/19.58$ & $2.6/2.8$ \\
BFLO & $1.79/0.65$ & $\mathbf{37.03}/\mathbf{37.69}$ & $16.92/16.67$
  & $62.76/62.35$ & $17.00/\mathbf{16.69}$ &
$\mathbf{1.8}/\mathbf{1.6}$\\
\bottomrule
\end{tabular}
\end{table*}

We used grid search to choose simple experimental
parameters that gave good results for SGD, and used them on all
datasets and gradient-based algorithms to isolate the effect of regularization. 
In particular, we avoided using popular
tricks such as ``heavy ball''/momentum \cite{Polyak1964}, minibatches, and
scheduling of the learning rate. SGD and BLANG were initialized with parameters
drawn from $N(0,\sigma^2)$ and BFLO with a prior equal to $N(0,\sigma^2)$, where
$\sigma = 0.2$. The learning rate was kept fixed at $\eta=0.001$. For the AROW
parameter we chose $r=10$. With the exception of EEG, all datasets were shuffled
at the beginning of a run. 

Table~\ref{tab:results_logistic} summarizes our experimental results
averaged over 10~runs. The online classification error is given within
a standard error $\sigma_\text{err} < 0.01$. The last column lists the
mean rank (out of~4, over all datasets), with~1 indicating an
algorithm attaining the best classification performance. BFLO falls
short in convergence speed due to its exploratory behavior in the
beginning, but then outperforms the other classifiers in its
generalization ability. Compared to the other methods, it comes
closest to the performance of BLANG, both in the transient and in the
steady-state regime, and in the remarkable robustness to noise.  This
may be because both methods use Monte Carlo samples of the posterior
to generate predictions.

\subsection{Feedforward Neural Networks}

This experiment investigates the application of Belief Flows to learn
the parameters of a more complex learning machine---in this case, a feedforward
neural network with one hidden layer. We compared the online performance of
Gaussian belief flows (BFLO) to two learning methods: plain SGD and SGD
with dropout \cite{Hinton2012}. As before, our aim was to isolate the 
effect of the regularization methods on the online
and test performance by choosing a simple experimental setup with shared global
parameters, avoiding architecture-specific optimization tricks.

We tested the learning algorithms on the well-known
MNIST\footnote{http://yann.lecun.com/exdb/mnist/index.html}
handwritten digit recognition task (abbreviated here as {\bf BASIC}), plus two
variations derived in \cite{Larochelle2007}: {\bf RANDOM}: MNIST digits
with a random background, where each random pixel value was drawn
uniformly; {\bf IMAGES}: a patch from a black and white image was used as the
background for the digit image. All datasets contained 62,000
grayscale, $28 \times 28$-pixel images (totalling 784 features). We split
each dataset into an online training set containing 80\% and a test set having
20\% of the examples.

\begin{table}
\newcommand{\kk}{\mathrm{K}}
\centering
\caption{MNIST Classification Results}
\label{tab:results_nn}
\scriptsize
\begin{tabular}{lcccc}
\toprule
& \multicolumn{4}{c}{Online Classification Error in \%} \\
\midrule
& PLAIN & RANDOM & IMAGES & Rank \\
$\max\{\sigma_\mathrm{err}\}$ & $0.07$ & $0.96$ & $1.16$ \\
\midrule
SGD & $11.25$ & $89.14$ & $72.41$ & $3.0$ \\
DROPOUT & $9.84$ & $52.87$ & $50.68$ & $1.6$ \\
BFLO & $11.01$ & $37.94$ & $47.71$ & $\mathbf{1.3}$ \\
\midrule
& \multicolumn{4}{c}{Final Classification Error in \%} \\
\midrule
& PLAIN & RANDOM & IMAGES & Rank \\
$\max\{\sigma_\mathrm{err}\}$ & $0.44$ & $3.33$ & $6.05$ \\
\midrule
SGD & $7.01$ & $89.17$ & $65.17$ & $3.0$ \\
DROPOUT & $5.52$ & $53.42$ & $46.67$ & $2.0$ \\
BFLO & $5.00$ & $29.11$ & $41.55$ & $\mathbf{1.0}$ \\
\bottomrule
\end{tabular}
\end{table}

To attain converging learning curves in the (single-pass) online setting, we
have chosen a modest architecture of 784 inputs, 200 hidden units and 10
outputs with aggressive updates. All units have a logistic sigmoid activation
function, and error gradients were evaluated on the binary Kullback-Leibler
divergence function averaged over the outputs. At the beginning of each run, the
examples were shuffled and the training initialized with weights either drawn
independently from a normal $N(0,\sigma^2)$ (SGD and dropout) or by setting the
prior to $N(0,\sigma^2)$ (BFLO), where $\sigma=0.1$. Throughout the online
learning phase, the learning rate was kept fixed at $\eta=0.2$ and we applied~5
update iterations on each example before discarding it. 

We report the results in Table~\ref{tab:results_nn} which were averaged over~5
runs. As the level of noise increased, all the classifiers declined in
performance, with PLAIN being the easiest and RANDOM being the hardest to
learn. SGD had a particularly poor behavior relative to the regularized
learners with their built-in mechanisms to avoid local minima. BFLO attained the
lowest error rates both online and in generalization. Interestingly, it copes
better with the RANDOM than with the IMAGES dataset. This could be because Monte
Carlo-sampling smoothens the gradients
within the neighborhood.

\section{Discussion and Future Work}

Our experiments indicate that Belief Flows methods are suited for
difficult online learning problems where robustness is a concern.
Gaussian belief flows may be related to ensemble learning methods in
conjunction with locally quadratic cost functions. A continous
ensemble of predictors that is trained under gradient descent follows
a linear velocity field when the objective function is a quadratic
form. Under such flow fields, the Gaussian family is a natural choice
for modelling the ensemble density since it is invariant to linear
velocity fields. An iteration of the update rule then infers the
dynamics of the whole ensemble from a single sample by conservatively
estimating the ensemble motion in terms of the Kullback-Leibler
divergence.

Following the Bayesian rationale, the resulting predictor is an
ensemble rather than a single member. Any strategy deciding which one
of them to use in a given round must deal with the
exploration-exploitation dilemma; that is, striking a compromise
between minimizing the prediction error and trying out new predictions
to further increase knowledge \cite{Sutton1998}.  This is necessary to
avoid local minima--here we sample a predictor according to the
probability of it being the optimal one, a strategy known as Thompson
sampling \cite{Thompson1933,Strens2000,OrtegaBraun2010e}.  The
maintainence of a dynamic belief in this manner allows the method to
outperform algorithms that maintain only a single estimate in complex
learning scenarios.

The basic scheme presented here can be extended in many ways. Some
possibilities include the application of Gaussian belief flows in
conjunction with kernel functions and gradient acceleration
techniques, and in closed-loop setups such as in active
learning. Furthermore, linear flow fields can be generalized by using
more complex probabilistic models suitable for other classes of flow
fields following the same information-theoretic framework outlined in
our work.  Finally, future theoretical work includes analyzing regret
bounds, and a more in-depth investigation of the relation between
stochastic approximation, reinforcement learning and Bayesian methods
that are synthesized in a belief flow model.

\comment{
\subsubsection*{Acknowledgements}
We thank the anonymous reviewers for their valuable comments and suggestions for
improving this manuscript. This study was funded by grants from the
U.S. National Science Foundation, Office of Naval Research and
Department of Transportation.
}

\appendices

\section{Proof of Theorem~1}

\begin{proof}
The KL divergence is given by:
\[
{\rm D}_{KL}(P_{n+1}\vert\vert P_{n})=\int
P_{n+1}(w)\log\frac{P_{n+1}(w)}{P_{n}(w)}dw.
\]
Then for Gaussians in $\mathbb{R}^{d}$, this divergence is:
\begin{align*}
\frac{1}{2}&(\mu_{n+1}-\mu_{n})^{T}\Sigma_{n}^{-1}(\mu_{n+1}-\mu_{n})
  +\frac{1}{2}{\rm Tr}\Sigma_{n}^{-1}\Sigma_{n+1} \\
  &-\frac{1}{2}\log\det\Sigma_{n}^{-1}\Sigma_{n+1}
  -\frac{d}{2}.
\end{align*}

The constraint on the flow implies:
\[
b=w_{n}^{\prime}-Aw_{n}
\]
\[
\mu_{n+1}-\mu_{n}=(w_{n}^{\prime}-\mu_{n})-A(w_{n}-\mu_{n})=\Delta_{n}^{\prime}
-A\Delta_{n}
\]
where $\Delta_{n}=w_{n}-\mu_{n}$ and
$\Delta_{n}^{\prime}=w_{n}^{\prime}-\mu_{n}$.
So the optimization can be written in terms of the matrix $A$:
\begin{align*}
\min_{A} &\frac{1}{2}\left[\Delta_{n}^{\prime} - A\Delta_{n}\right]^{T}
\Sigma_{n}^{-1} \left[\Delta_{n}^{\prime}-A\Delta_{n}\right] \\
&+ \frac{1}{2}{\rm Tr}\Sigma_{n}^{-1} A\Sigma_{n}A^{T}
- \frac{1}{2}\log\det AA^{T}
- \frac{d}{2}.
\end{align*}

We first note that minimizing the KL-divergence implies that the transformation
$A$ is full-rank, since collapsing the rank of the covariance matrix
will lead to infinite divergence. Taking the derivative with respect
to $A$ yields
\[
-\Sigma_{n}^{-1}\left[\Delta_{n}^{\prime}
-A\Delta_{n}\right]\Delta_{n}^{T}
-\Sigma_{n}^{-1}A\Sigma_{n}-A^{-T} = 0,
\]
which can be rewritten as
\begin{align*}
\Sigma_{n}
&=A\Sigma_{n}A^{T}-(\Delta_{n}^{\prime}-A\Delta_{n})(A\Delta_{n})^{T}
\nonumber \\
&=A\left(\Sigma_{n}
  +\Delta_{n} \Delta_{n}^{T}\right)A^{T}
  -\Delta_{n}^{\prime}(A\Delta_{n})^{T}.
  \label{eq:full_derivative}
\end{align*}
We first see that if $\Delta_{n}^{\prime}=\Delta_{n}$, then $A=I$
is the solution where the flow field is invariant. The expression
can be simplified if we consider the diagonalization of the covariance
matrix:
\[
\Sigma_{n}=U_{n}D_{n}U_{n}^{T}.
\]
In that case, we transform the variables:
\begin{gather*}
\tilde{\Delta}_{n}
  \leftarrow\frac{1}{\sqrt{D_{n}}}U_{n}^{T}\Delta_{n} \tilde{\Delta}_{n}^{\prime}
  \leftarrow\frac{1}{\sqrt{D_{n}}}U_{n}^{T}\Delta_{n}^{\prime} \tilde{A}
  \\
  \leftarrow\frac{1}{\sqrt{D_{n}}}U_{n}^{T}AU_{n}\sqrt{D_{n}}.
\end{gather*}

So, in terms of these transformed variables, the optimal condition
becomes:
\begin{equation}\label{eq:full_derivative_transformation}
I=\tilde{A}(I+\tilde{\Delta}_{n}\tilde{\Delta}_{n}^{T})\tilde{A}^{T}-\tilde{
\Delta}_{n}^{\prime}(\tilde{A}\tilde{\Delta}_{n})^{T}.
\end{equation}

The general solution can be found by considering the 2-D basis spanned by the
vectors $\tilde{\Delta}_{n}$ and $\tilde{\Delta}_{n}^{\prime}$, with unit
vectors $\hat{\mu}$ and $\hat{\nu}$.
\begin{align}
\tilde{\Delta}_{n} &= u\hat{\mu}
& \tilde{\Delta}_{n}^{\prime} &= v_{\parallel}\hat{\mu}+v_{\perp}\hat{\nu}.
\end{align}
The optimal matrix $\tilde{A}$ is just identity in the other
$D-2$ directions orthogonal to this 2-D subspace. Then the optimality
condition in \eqref{eq:full_derivative_transformation} can be written
in terms of the $2\times2$ matrix
\begin{equation}
A_{2\times2}=\left[\begin{array}{cc}
A_{\mu\mu} & A_{\mu\nu}\\
A_{\nu\mu} & A_{\nu\nu}
\end{array}\right]
\end{equation}
as 
\begin{equation}
A_{2\times2}\left[\begin{array}{cc}
1+u^{2} & 0\\
0 & 1
\end{array}\right]A_{2\times2}^{T}-\left[\begin{array}{cc}
v_{\parallel}u & 0\\
v_{\perp}u & 0
\end{array}\right]A_{2\times2}^{T}=I\label{eq:2x2condition}.
\end{equation}

To solve this quadratic matrix equation, we first note that symmetry
of the matrices implies that the solution must satisfy:
$\Delta_{n}^{\prime}\Vert(A\Delta_{n})$.
This means that the solution must be of the restricted form:
\[
A_{2\times2}=\left[\begin{array}{cc}
\alpha v_{\parallel} & A_{\mu\nu}\\
\alpha v_{\perp} & A_{\nu\nu}
\end{array}\right]
\]

Then in terms of the unknowns $\alpha$, $A_{\mu\nu}$, and $A_{\nu\nu}$,
we have the following three equations:
\begin{align*}
1 &= v_{\parallel}^{2}\left[(1+u^{2})\alpha^{2}
  - u\alpha\right]+A_{\mu\nu}^{2} \\
1 &= v_{\perp}^{2}\left[(1+u^{2})\alpha^{2}
  - u\alpha\right]+A_{\nu\nu}^{2} \\
0 &=
v_{\parallel}v_{\perp}\left[(1+u^{2})\alpha^{2}-u\alpha\right]
  + A_{\mu\nu}A_{\nu\nu}.
\end{align*}

Fortunately, we can determine the solution analytically in closed
form:
\begin{gather*}
A_{\mu\nu}
  =\mp\frac{v_{\perp}}{\sqrt{v_{\parallel}^{2}+v_{\perp}^{2}}}
\qquad A_{\nu\nu}
  =\pm\frac{v_{\parallel}}{\sqrt{v_{\parallel}^{2}+v_{\perp}^{2}}} \\
(1+u^{2})\alpha^{2}-u\alpha
  =\frac{1}{v_{\parallel}^{2}+v_{\perp}^{2}}.
\end{gather*}

The quadratic formula then yields:
\[
\alpha =
\frac{ u \pm \sqrt{u^{2}
  + \frac{4(1+u^{2})}{v_{\parallel}^{2} + v_{\perp}^{2}}}
}{2(1+u^{2})}
\]
Putting these together, we get that the optimal transformation matrix
is:
\[
A_{2\times2} =
\frac{1}{\sqrt{v_{\parallel}^{2}+v_{\perp}^{2}}}
\left[\begin{array} {cc}
\frac{u\sqrt{v_{\parallel}^{2}+v_{\perp}^{2}}
  \pm \sqrt{4+u^{2}(4+v_{\parallel}^{2}
  + v_{\perp}^{2})}}{2(1+u^{2})}v_{\parallel}
& \mp v_{\perp} \\
\frac{u\sqrt{v_{\parallel}^{2} + v_{\perp}^{2}}
  \pm \sqrt{4+u^{2}(4+v_{\parallel}^{2}
  + v_{\perp}^{2})}}{2(1+u^{2})}v_{\perp}
& \pm v_{\parallel}
\end{array}\right]
\]

We see that there are actually four discrete solutions for $A_{2\times2}$.
First, one of two roots (one positive, one negative) for $\alpha$ can be
chosen, and then the signs for $A_{\mu\nu}$ and $A_{\nu\nu}$ can
be swapped. However, if we consider the situation where $\Delta_{n}^{\prime}$
is not far from $\Delta_{n}$, then we should choose the solution
connected to the identity matrix $I$. This means selecting the positive
root for $\alpha$, and ensuring the diagonal terms of $A_{2\times2}$
are positive. The full matrix solution $A^{\ast}$ can be expressed
in terms of this $A_{2\times2}$ as:
\begin{equation}\label{eq:transformation}
A^{\ast} =
I_{D\times D} \\
+ U_{n} \sqrt{D_{n}} M \frac{1}{\sqrt{D_{n}}} U_{n}^{T}
\end{equation}
where
\[
  M = \left\{
  \left[\begin{array}{cc}
  \hat{\mu} & \hat{\nu}
  \end{array}\right]
  \left(A_{2\times2} - I_{2\times2}\right)
  \left[\begin{array}{c}
  \hat{\mu}^{T}\\
  \hat{\nu}^{T}
  \end{array}\right]
  \right\}.
\]

The parameters for the new belief distribution in the next round are
then given in terms of $A^{\ast}$:
\begin{equation}\label{eq:update_full}
\Sigma_{n+1}=A^{\ast}\Sigma_{n}A^{\ast T}
\qquad
\mu_{n+1}=A^{\ast}(\mu_{n}-w_{n})+w_{n}^{\prime}.
\end{equation}
This concludes our proof.
\end{proof}

\subsection{Special Cases}

The optimal solution for the diagonal flow is obtained as a special case of
\eqref{eq:2x2condition} where $\Delta_{n}^{\prime}\parallel\Delta_{n}$ and
$v_{\perp}=0$. Let $\sigma_{i}^{2}$ be the variance of $\Sigma_{n}$ along the
$i$-th component. If we rescale the parameter components according to
$\sigma_i^2$ as
\begin{align*}
u_{i} &= \frac{w_{i}-\mu_{i}}{\sigma_{i}}
& v_{i} &= \frac{w_{i}^{\prime}-\mu_{i}}{\sigma_{i}},
\end{align*}
then the optimal scaling parameter is given by
\[
a_{i}=\frac{u_{i}v_{i}\pm\sqrt{4+u_{i}^{2}(4+v_{i}^{2})}}{2(1+u_{i}^{2})}.
\]
For the spherical case, we begin by noting that the covariance and the
transformation matrices are all isotropic, i.e.\ of the form $M = m U$, where
$m$ is a scalar and $U$ is unitary. Because of this, the multivariate problem
effectively reduces to a univariate problem where $A = a U$ first rotates
$(w-\mu)$ to align it to $(w'-\mu)$ and then scales and translates the
distribution along this axis.

\bibliographystyle{IEEEtran.bst}
\bibliography{bibliography}
\end{document}